\documentclass[11pt]{article}

\usepackage{fullpage}
\usepackage[ruled, linesnumbered, vlined, commentsnumbered]{algorithm2e}
\usepackage{graphicx,subfig} 
\usepackage{amsfonts,amssymb,amsmath,amsthm,amsopn,mathtools}	
\usepackage{booktabs,diagbox,colortbl,multirow,tabularx,threeparttable,hhline}
\usepackage[listings,skins,breakable]{tcolorbox}

\usepackage{enumerate}
\usepackage{authblk}
\usepackage{footnote}
\usepackage{hyperref}
\usepackage{prettyref}
\usepackage{cite}
\usepackage{setspace}
\usepackage{color}
\usepackage{xcolor}  
\usepackage{scrpage2}
\usepackage{geometry}

\usepackage{tikz}
\usepackage{pgfplots}
\usetikzlibrary{positioning,shapes,shadows,arrows,calc}
\tikzstyle{component}=[rectangle, draw=black, rounded corners, fill=blue!40, drop shadow, text centered, anchor=north, text=white, minimum height=1cm]
\tikzstyle{arrow}=[->, thick]

\pgfplotsset{compat=1.12}
\usetikzlibrary{intersections}
\usetikzlibrary{pgfplots.statistics}
\usepgfplotslibrary{fillbetween}

\definecolor{myblue}{RGB}{34,31,217}
\definecolor{mycyan}{gray}{.7}
\definecolor{Gray}{gray}{0.9}

\newtheorem{remark}{Remark}
\newtheorem{theorem}{Theorem}

\newtheorem{corollary}{Corollary}
\newtheorem{definition}{Definition}

\newcommand{\our}{\texttt{D$^2$EMO/MGD}}

\DeclareMathOperator*{\argmax}{argmax}
\DeclareMathOperator*{\argmin}{argmin}

\newcommand{\bb}[1]{\multicolumn{1}{>{\columncolor{mycyan}}c}{\textbf{{#1}}}}

\newcommand{\pref}{\prettyref}

\newrefformat{fig}{Fig.~\ref{#1}}
\newrefformat{tab}{Table~\ref{#1}}
\newrefformat{sec}{Section~\ref{#1}}
\newrefformat{alg}{Algorithm~\ref{#1}}
\newrefformat{property}{Property~\ref{#1}}
\newrefformat{theorem}{Theorem~\ref{#1}}
\newrefformat{definition}{Definition~\ref{#1}}
\newrefformat{corollary}{Corollary~\ref{#1}}
\newrefformat{lemma}{Lemma~\ref{#1}}
\newrefformat{conj}{Conjecture~\ref{#1}}
\newrefformat{def}{Definition~\ref{#1}}
\newrefformat{eq}{equation~(\ref{#1})}
\newrefformat{app}{Appendix~\ref{#1}}

\usepackage{lscape}

\begin{document}

\title{\vspace{-1ex}\LARGE\textbf{Data-Driven Evolutionary Multi-Objective Optimization Based on Multiple-Gradient Descent for Disconnected Pareto Fronts}\footnote{Both authors made equal contributions to this paper.}~\footnote{This manuscript is submitted for potential publication. Reviewers can use this version in peer review.}}

\author[1]{\normalsize Renzhi Chen}
\author[2]{\normalsize Ke Li}
\affil[1]{\normalsize PLA Academy of Military Science, Beijing, China}
\affil[2]{\normalsize Department of Computer Science, University of Exeter, EX4 4QF, Exeter, UK}
\affil[$\ast$]{\normalsize Email: \texttt{k.li@exeter.ac.uk}}

\date{}
\maketitle

\vspace{-3ex}
{\normalsize\textbf{Abstract: } }Data-driven evolutionary multi-objective optimization (EMO) has been recognized as an effective approach for multi-objective optimization problems with expensive objective functions. The current research is mainly developed for problems with a \lq regular\rq\ triangle-like Pareto-optimal front (PF), whereas the performance can significantly deteriorate when the PF consists of disconnected segments. Furthermore, the offspring reproduction in the current data-driven EMO does not fully leverage the latent information of the surrogate model. Bearing these considerations in mind, this paper proposes a data-driven EMO algorithm based on multiple-gradient descent. By leveraging the regularity information provided by the up-to-date surrogate model, it is able to progressively probe a set of well distributed candidate solutions with a convergence guarantee. In addition, its infill criterion recommends a batch of promising candidate solutions to conduct expensive objective function evaluations. Experiments on $33$ benchmark test problem instances with disconnected PFs fully demonstrate the effectiveness of our proposed method against four selected peer algorithms.

{\normalsize\textbf{Keywords: } }Data-driven optimization  \and Multiple-gradient descent \and Evolutionary multi-objective optimization.


\section{Introduction}
\label{sec:introduction}

Many real-world scientific and engineering applications involve multiple conflicting objectives, a.k.a. multi-objective optimization problems (MOPs). For example, tuning a water distribution system to optimize its financial and operational costs~\cite{MarquesCS15}, minimizing the energy consumption while maximizing locomotion speed in a complex robotic system~\cite{AriizumiTCM14}. In multi-objective optimization, there does not exist a solution that optimizes all conflicting objectives simultaneously. Instead, we are looking for a set of representative, with a promising convergence and diversity, trade-off solutions that compromise one objective for another.

Due to the population-based characteristics, evolutionary algorithms (EAs) have been widely recognized as an effective approach for MO~\cite{Deb01,LaiL021}. However, one of the major criticisms of EAs is its daunting amount of function evaluations (FEs) required to obtain a set of reasonable solutions. This is unfortunately unacceptable in practice since FEs are either computationally or financially demanding, e.g., computational fluid dynamic simulations can take from minutes to hours to carry out a single FE~\cite{JinS09}. To mitigate this issue, data-driven evolutionary optimization\footnote{It is also known as surrogate-assisted EA interchangeably in the literature~\cite{Jin05}.}, guided by surrogate models of computationally expensive objective functions, have become as a powerful approach for solving expensive optimization problems~\cite{JinWCGM19}. For example, some researchers considered various ways to build a surrogate model of the expensive objective functions, either collectively~\cite{LoshchilovSS10,AkhtarS16,ChughJMHS18,HabibSCRM19} or as a weighted aggregation~\cite{Knowles06,ZhangLTV10,SunLGHL11,MartinezC13}. According to the ways of surrogate modeling, bespoke model management strategies are developed to select promising candidate solution(s) for conducting expensive FEs. In particular, this can either be driven by the surrogate model directly~\cite{SunLGHL11,MartinezC13,AkhtarS16} or an acquisition function inferred from the model uncertainty~\cite{Knowles06,ZhangLTV10,LoshchilovSS10,ChughJMHS18,HabibSCRM19}. There are two gaps in the current literature that hinder the further uptake of data-driven evolutionary multi-objective optimization (EMO) in practice.
\begin{itemize}
    \item Most, if not all, existing studies are mainly designed and validated on prevalent test problems (e.g., DTLZ1 to DTLZ4~\cite{DebTLZ05} and WFG4 to WFG9~\cite{HubandHBW06}) characterized as \lq regular\rq\ triangle-like Pareto-optimal fronts (PFs). Unfortunately, this is unrealistic in the real-world optimization scenarios~\cite{IshibuchiHS19}. On the contrary, it is not uncommon that the PFs of real-world applications are featured as disconnected, incomplete, degenerated, and/or badly-scaled (partially due to the complex and nonlinear relationship between objectives), it is surprising that the research on handling MOPs with irregular PFs is lukewarm in the context of data-driven EMO, except for~\cite{HabibSCRM19}.
    \item In addition, the evolutionary operators for offspring reproduction are directly derived from the EA (e.g., crossover and mutation~\cite{Holland73}, differential evolution~\cite{StornP97}, particle swarm optimization~\cite{KennedyE95}) or conventional mathematical programming (e.g., simplex~\cite{DicksonF60}, Nelder–Mead~\cite{NelderM65}, and trust-region methods~\cite{Powell84}). By this means, the regularity information of the underlying MOP embedded in the surrogate model(s) is unfortunately yet exploited. Note that such information can be beneficial to navigate a more effective exploration of the search space.
\end{itemize}

Bearing these considerations in mind, this paper proposes a data-driven evolutionary multi-objective optimization based on multiple-gradient descent~\cite{Desideri14} for expensive MOPs with disconnected PFs. Its basic idea is to leverage the gradient information of the surrogate models to explore promising candidate solutions. It consists of the following two distinctive components.
\begin{itemize}
    \item \texttt{MGD-based evolutionary search}: As the main crux of our proposed algorithm, it generates a set of candidate solutions guided by the multiple-gradient descent of the surrogate model of each computationally expensive objective function. In a nutshell, these candidate solutions are first randomly sampled in the decision space. Then, they are gradually guided to interpolate well distributed potential solutions along the manifold of the surrogate PS.
    \item \texttt{Infill criterion}: It recommends a batch of promising candidate solutions obtained by the \texttt{MGD-based evolutionary search} step to carry out expensive FEs for the model management.
\end{itemize}
Our experiments on $33$ benchmark test problem instances with disconnected PFs fully demonstrate the effectiveness and outstanding performance of our proposed \our\ against four selected peer algorithms.

The rest of this paper is organized as follows. \pref{sec:preliminaries} gives some preliminary knowledge pertinent to this paper. The technical details of our proposed method is introduced in~\pref{sec:method}. The experimental setup is given in~\pref{sec:setup} and the results are presented and discussed in~\pref{sec:results}. \pref{sec:conclusions} concludes this paper and sheds some lights on potential future directions.

\section{Preliminaries}
\label{sec:preliminaries}

In this section, we give some basic definitions pertinent to this paper.

\subsection{Basic Definitions in Multi-Objective Optimization}
\label{sec:definitions}

The MOP considered in this paper is defined as:
\begin{equation}
    \begin{array}{l l}
        \mathrm{minimize} \quad \mathbf{F}(\mathbf{x})=(f_{1}(\mathbf{x}),\cdots,f_{m}(\mathbf{x}))^{T}\\
        \mathrm{subject\ to} \;\; \mathbf{x} \in\Omega
    \end{array},
    \label{eq:MOP}
\end{equation}
where $\mathbf{x}=(x_1,\cdots,x_n)^T$ is a decision vector and $\mathbf{F}(\mathbf{x})$ is an objective vector. $\Omega=[x_i^L,x_i^U]^n_{i=1}\subseteq\mathbb{R}^n$ defines the search space. $\mathbf{F}: \Omega\rightarrow\mathbb{R}^m$ is the corresponding attainable set in the objective space $\mathbb{R}^m$.
\begin{definition}
\label{def:dominate}
    Given two solutions $\mathbf{x}^1,\mathbf{x}^2\in\Omega$, $\mathbf{x}^1$ is said to \underline{Pareto dominate} $\mathbf{x}^2$, denoted as $\mathbf{x}^1\preceq\mathbf{x}^2$, if and only if $f_i(\mathbf{x}^1)\leq f_i(\mathbf{x}^2)$ for all $i\in\{1,\cdots,m\}$ and $\mathbf{F}(\mathbf{x}^1)\neq\mathbf{F}(\mathbf{x}^2)$. 
\end{definition}
\begin{definition}
    A solution $\mathbf{x}^\ast\in\Omega$ is said to be \underline{Pareto-optimal} if and only if $\nexists\mathbf{x}^\prime\in\Omega$ such that $\mathbf{x}^\prime\preceq\mathbf{x}^\ast$.
\end{definition}
\begin{definition}
    The set of all Pareto-optimal solutions is called the \underline{Pareto-optimal set} (PS), i.e., $\mathcal{PS}=\{\mathbf{x}^\ast|\nexists\mathbf{x}^\prime\in\Omega \text{ such that } \mathbf{x}^\prime\preceq\mathbf{x}^\ast\}$ and their corresponding objective vectors form the \underline{Pareto-optimal front} (PF), i.e., $\mathcal{PF}=\{\mathbf{F}(\mathbf{x}^\ast)|\mathbf{x}^\ast\in\mathcal{PS}\}$.
\end{definition}

\subsection{Gaussian Process Regression Model}
\label{sec:GP}

In view of the continuously differentiable property, we consider the Gaussian process regression (GPR)~\cite{RasmussenW06} as the surrogate model of each expensive objective function. Given a set of training data $\mathcal{D}=\{(\mathbf{x}^i,f(\mathbf{x}^i)\}_{i=1}^{N}$, a GPR model aims to learn a latent function $g(\mathbf{x})$ by assuming $f(\mathbf{x}^i)=g(\mathbf{x}^i)+\epsilon$ where $\epsilon\sim\mathcal{N}(0,\sigma^2_n)$ is an independently and identically distributed Gaussian noise. For each testing input vector $\mathbf{z}^\ast\in\Omega$, the mean and variance of the target $f(\mathbf{z}^\ast)$ are predicted as:
\begin{align}
\overline{g}(\mathbf{z}^\ast)&=m(\mathbf{z}^\ast)+{\mathbf{k}^\ast}^T(K+\sigma_n^2 I)^{-1}(\mathbf{f}-\mathbf{m}(X)),\\
\mathbb{V}[g(\mathbf{z}^\ast)]&=k(\mathbf{z}^\ast,\mathbf{z}^\ast)-{\mathbf{k}^\ast}^T (K+\sigma_n^2 I)^{-1} {\mathbf{k}^\ast},
\label{eq:GP}
\end{align}
where $X=(\mathbf{x}^1,\cdots,\mathbf{x}^N)^T$ and $\mathbf{f}=(f(\mathbf{x}^1),\cdots,f(\mathbf{x}^N))^T$. $\mathbf{m}(X)$ is the mean vector of $X$, $\mathbf{k}^\ast$ is the covariance vector between $X$ and $\mathbf{z}^\ast$, and $K$ is the covariance matrix of $X$. In this paper, we use the radial basis function as the covariance function to measure the similarity between a pair of two solutions $\mathbf{x}$ and $\mathbf{x}^\prime\in\Omega$:
\begin{equation}
    k(\mathbf{x},\mathbf{x}^\prime)=\gamma\exp(-\frac{\|\mathbf{x}-\mathbf{x}^\prime\|^2}{\ell}),
\end{equation}
where $\|\cdot\|$ is the Euclidean norm and $\gamma$ and length scale $\ell$ are two hyperparameters. The predicted mean $\overline{g}(\mathbf{z}^\ast)$ is directly used as the prediction of $f(\mathbf{z}^\ast)$, and the predicted variance $\mathbb{V}[g(\mathbf{x}^\ast)]$ quantifies the uncertainty. In practice, the hyperparameters associated with the mean and covariance functions are learned by maximizing the log marginal likelihood function as recommended in~\cite{RasmussenW06}. 
For the sake of simplicity, here we assume that the mean function is a constant $0$ and the inputs are noiseless.


\section{Proposed Method}
\label{sec:method}

\begin{figure}[t!]
    \centering
    \includegraphics[width=\linewidth]{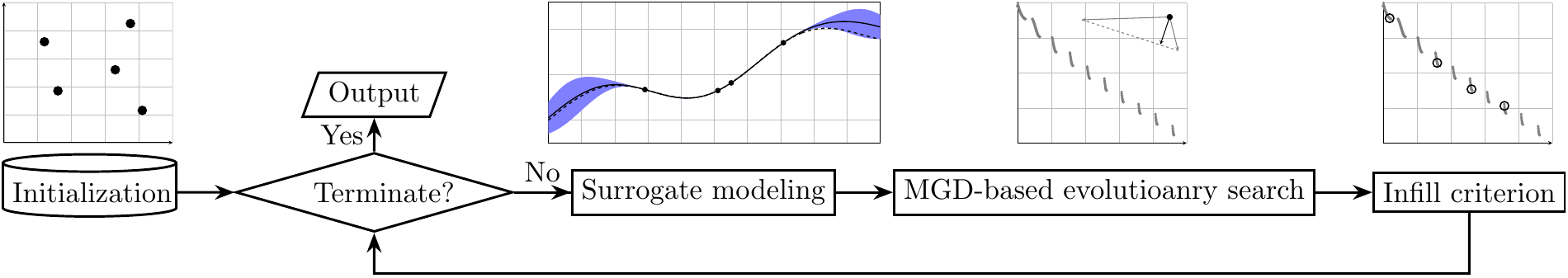}
    \caption{The flow chart of the proposed \our.}
    \label{fig:flowchart}
\end{figure}

In this section, we plan to delineate the implementation of our proposed data-driven evolutionary multi-objective optimization based on multiple-gradient descent (dubbed \our). As the flowchart shown in~\pref{fig:flowchart}, \our\ starts with an \texttt{initialization} step based on an experimental design method such as Latin hypercube sampling~\cite{SantnerWN18}. Note that these initial samples will be evaluated based on the computationally expensive objective functions. During the main loop, the \texttt{surrogate modeling} step builds a surrogate model by using the GPR for each expensive objective function based on the data collected so far. The other two steps will be delineated in the following paragraphs.

\subsection{MGD-based Evolutionary Search}
\label{sec:mgd_search}

This step aims to search for a set of promising candidate solutions $\mathcal{P}=\{\hat{\mathbf{x}}^i\}_{i=1}^{\tilde{N}}$, which are assumed to be an appropriate approximation to the PF, based on the surrogate model built in the \texttt{surrogate modeling} step. The working mechanism of this \texttt{MGD-based evolutionary search} step is given as follows.
\begin{enumerate}[Step 1:]
    \item Initialize a candidate solution set
        $\mathcal{P}=\{\hat{\mathbf{x}}^i\}_{i=1}^{\tilde{N}}$ based on Latin hypercube sampling upon $\Omega$.
    \item For each solution $\hat{\mathbf{x}}^i\in\mathcal{P}$, do
        \begin{enumerate}[Step 2.1:]
            \item Calculate the gradient of the predicted mean of each objective function $\nabla\overline{g}_j(\hat{\mathbf{x}}^i)$ where $j\in\{1,\cdots,m\}$.
            \item Find a nonnegative unit vector $\mathbf{w}^\ast=(w_1^\ast,\cdots,w_m^\ast)^\top$ that satisfies:
                \begin{equation}
                    \mathbf{w}^\ast=\argmin_{\mathbf{w}}\bigg\|\sum_{j=1}^m w_j\nabla\overline{g}_j(\hat{\mathbf{x}}^i)\bigg\|,
                    \label{eq:unit_weight}
                \end{equation}
                where $\mathbf{w}=(w_1,\cdots,w_m)^\top$, $\sum_{i=1}^m w_i=1$ and $w_i\geq 0$, $i\in\{1,\cdots,m\}$.
            \item Obtain a directional vector $\mathbf{u}^\ast$ as:
                \begin{equation}
                    \mathbf{u}^\ast=
                        \begin{cases}
                            \underset{1\leq j\leq m}{\argmax}\big\|\nabla\overline{g}_j(\hat{\mathbf{x}}^i)\big\|, & \text{if }\sum_{j=1}^m w_j^\ast\nabla\overline{g}_j(\hat{\mathbf{x}}^i)=0 \\
                            \underset{1\leq j\leq m}{\argmin}\big\|\nabla\overline{g}_j(\hat{\mathbf{x}}^i)\big\|, & \text{if }\exists i,j\in\{1,\cdots,m\},\langle\nabla\overline{g}_j(\hat{\mathbf{x}}^i),\nabla\overline{g}_j(\hat{\mathbf{x}}^i)\rangle>\delta\\
                            \sum_{j=1}^m w_j^\ast\nabla\overline{g}_j(\hat{\mathbf{x}}^i), & \text{otherwise} 
                        \end{cases}
                        \label{eq:unit_u}
                \end{equation}
                where $\langle\ast,\ast\rangle$ measures the acute angle between two vectors, and $\delta=\min\Big\{\big\|\nabla\overline{g}_k(\hat{\mathbf{x}}^i)\big\|\Big\}_{k=1}^m$.
            \item Amend the updated solution to $\mathcal{P}\leftarrow\mathcal{P}\bigcup\{\hat{\mathbf{x}}^i+\eta\mathbf{u}^\ast\}$
        \end{enumerate}
    \item Remove the dominated solutions in $\mathcal{P}$ according to their predicted objective functions.
    \item If the stopping criterion is met, then stop and output $\mathcal{P}$. Otherwise, go to Step 2.
\end{enumerate}

\begin{remark}
    As discussed in~\cite{Desideri14}, the multiple-gradient descent (MGD) is a natural extension of the single-objective gradient to finding a PF. In a nutshell, its basic idea is to iteratively update a solution $\mathbf{x}$ along a \lq specified\rq\ direction so that all objective functions can thus be improved. Different from the linear weighted aggregation, the MGD works for non-convex PF. Therefore, we can expect a descent diversity in case the initial population is well distributed. Note that since the objective functions are assumed to be as a black box a priori, the MGD is not directly applicable in our context.
\end{remark}

\begin{remark}
    In this paper, since the computationally expensive objective functions are modeled by GPR, which is continuously differentiable, we can derive the gradient of the predicted mean function w.r.t. a solution $\mathbf{x}$ as:
    \begin{equation}
        \frac{\partial\overline{g}(\mathbf{x})}{\partial \mathbf{x}}=\frac{\partial\mathbf{k}^\ast}{\partial\mathbf{x}}K^{-1}\mathbf{f},
    \end{equation}
    where the first-order derivative of $\mathbf{k}^\ast$, i.e., the covariance vector between $\mathcal{P}$ and $\mathbf{x}$, is calculated as:
    \begin{equation}
        \frac{\partial\mathbf{k}^\ast}{\partial\mathbf{x}}=-\frac{\partial\|\mathbf{x}-\mathbf{x}^\prime\|}{\partial\mathbf{x}}\frac{\overline{g}(\mathbf{x})}{\ell}.
    \end{equation}
\end{remark}

\begin{remark}
    The optimization problem in (\ref{eq:unit_weight}) is essentially equivalent to finding a minimum-norm point in the convex hull. When $m=2$, we have the closed form solution as:
    \begin{equation}
        \label{eq:cal_w_2d}
        w_1^\ast = 
        \frac{\big(\nabla\overline{g}_2(\hat{\mathbf{x}}^i)-\nabla\overline{g}_1(\hat{\mathbf{x}}^i)\big)^\top\nabla\overline{g}_2(\hat{\mathbf{x}}^i)}{\|\nabla\overline{g}_2(\hat{\mathbf{x}}^i)-\nabla\overline{g}_1(\hat{\mathbf{x}}^i)\big\|^2}, \text{ } w_2^\ast=1-w_1^\ast.
    \end{equation}
\end{remark}

\begin{figure}[t!]
    \centering
    \includegraphics[width=\linewidth]{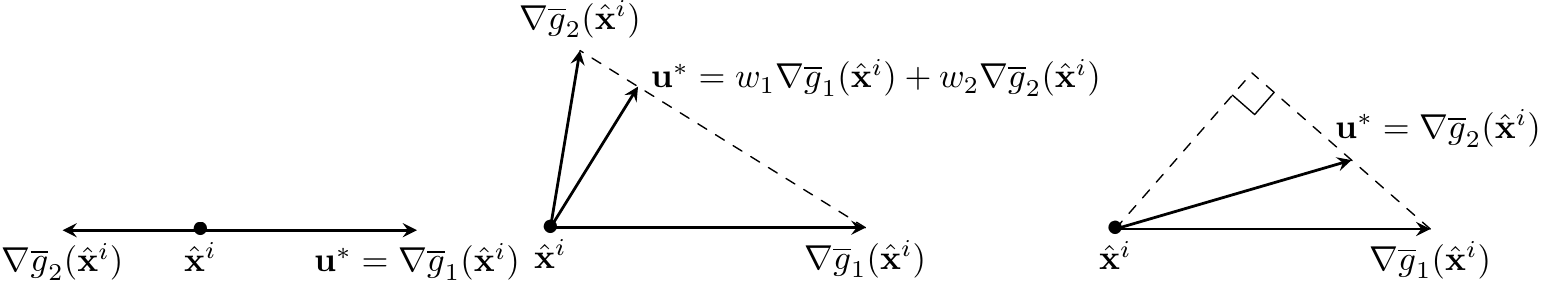}
    \caption{Illustrative examples of the calculation of $\mathbf{u}^\ast$ in~\pref{eq:unit_u}.}
    \label{fig:example_u}
\end{figure}

\begin{remark}
    \pref{fig:example_u} gives an illustrative example for each of the three conditions given in~\pref{eq:unit_u} when $m=2$. More specifically, when the gradients of two objective functions are in opposite directions as shown in~\pref{fig:example_u}(a), $\mathbf{u}^\ast$ is chosen as the one with a larger Euclidean norm. If the gradients are too close to each other as shown in~\pref{fig:example_u}(b), $\mathbf{u}^\ast$ is chosen as the one with a smaller Euclidean norm. On the contrary, $\mathbf{u}^\ast$ is set as the weighted aggregation of two gradients where the weights are obtained from Step 2.2.
\end{remark}

\begin{remark}
    According to the Karush–Kuhn–Tucker (KKT) conditions~\cite{KuhnT51}, we have $\forall\mathbf{x}^\ast\in\mathcal{PS}$, $\exists\vec{\alpha}=(\alpha_1,\cdots,\alpha_m)^T$, where $\alpha_i\geq 0$, $i\in\{1,\cdots,m\}$ and $\sum_{i=1}^m\alpha_i=1$, such that $\sum_{i=1}^m\alpha_i\nabla f_i(\mathbf{x}^\ast)=0$. In this case, we come up with~\pref{theorem:theorem_convergent_direction} and~\pref{corollary:on_ps}, the proof of which can be found in the supplemental document of this paper\footnote{The supplemental document can be found from~\url{xxxxxxxxxxxxxxxxx}.}.
\end{remark}

\begin{theorem}
    Considering the $m$ objective functions defined in~(\ref{eq:MOP}), $\forall\mathbf{x}\in\Omega$ but $\mathbf{x}\notin\mathcal{PS}$, $\exists\mathbf{w}^\ast$ that satisfies~(\ref{eq:unit_weight}) and $\mathbf{u}^\ast=\sum_{j=1}^m w_j^\ast\nabla f_j(\mathbf{x})>0$, we can obtain a new solution $\mathbf{x}^\prime=\mathbf{x}+\eta\mathbf{u}^\ast$ such that $\mathbf{x}^\prime\preceq\mathbf{x}$.
    \label{theorem:theorem_convergent_direction}
\end{theorem}

\begin{corollary}
    Considering the $m$ objective functions defined in~(\ref{eq:MOP}), $\forall\mathbf{x}\in\mathcal{PS}$, $\exists\mathbf{w}^\ast$ that satisfies~(\ref{eq:unit_weight}) and $\mathbf{u}^\ast=\sum_{j=1}^m w_j^\ast\nabla f_j(\mathbf{x})=0$, we can obtain a new solution $\mathbf{x}^\prime=\mathbf{x}+\eta\mathbf{u}^\ast$ such that $\mathbf{x}^\prime$ is still on the PS.
    \label{corollary:on_ps}
\end{corollary}

\begin{figure}[t!]
    \centering
    \includegraphics[width=\linewidth]{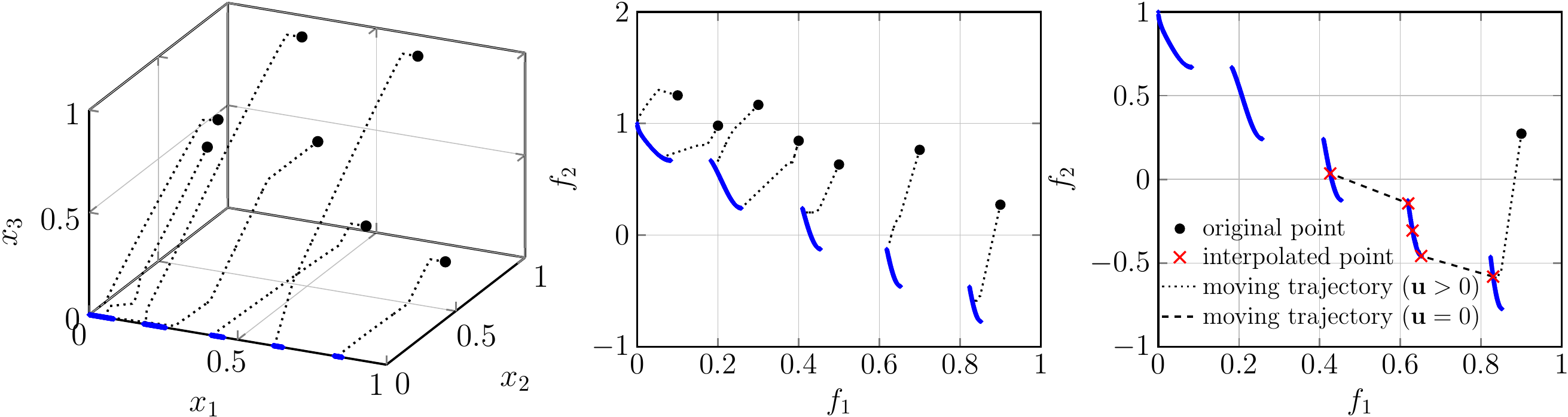}
    \caption{Illustrative examples of \texttt{MGD} w.r.t. different solutions moving towards different (a) PS segments and (b) PF segments, and (c) the working mechanismthe \texttt{MGD-based evolutionary search} step.}
    \label{fig:mgd_example}
\end{figure}

\begin{remark}
    According to~\pref{theorem:theorem_convergent_direction}, the \texttt{MGD-based evolutionary search} step can be understood as pushing a solution towards the PS first before implementing a random walk along the PS as an illustrative example shown in~\pref{fig:mgd_example}.
\end{remark}

\begin{remark}
    In Step 2.4, $\eta\in(0,1]$ is a random scaling factor along the direction vector $\mathbf{u}^\ast$. The stopping criterion in Step 4 is the number of iterations (here it is set as $100$ in our experiments) of this \texttt{MGD-based evolutionary search} step.
\end{remark}

\subsection{Infill Criterion}
\label{sec:infill}

This step aims to pick up $\xi\geq 1$ promising solutions from $\mathcal{P}$ and evaluate them by using the computationally expensive objective functions. These newly evaluated solutions are then used to update the training dataset for the next iteration. In a nutshell, there are two main difference w.r.t. many existing works on data-driven EA~\cite{JinWCGM19} and Bayesian optimization~\cite{abs-1807-02811}. First, even though we the GPR as the surrogate model, our infill criterion does not rely on an uncertainty quantification measure, a.k.a. acquisition function. Second, instead of recommending one solution for the computationally expensive function evaluation in a sequential manner, the \texttt{infill criterion} step of \our\ proposes to select a batch of samples at a time. Under a limited computational budget, we can expect to reduce the number of iterations of the main loop in~\pref{fig:flowchart} by $\xi$ times. In addition, since many physical experiments can be carried out in parallel given the availability of more than one infrastructure (e.g., the training and validating machine learning models are usually distributed into multiple cores or GPUs for hyper-parameter optimization in automated machine learning), such batched recommendation provides an actionable way for parallelization. Therefore, we can anticipate the practical importance to save the computational overhead. In this paper, we propose a simple infill criterion based on the individual Hypervolume~\cite{ZitzlerT99} contribution (IHV). Specifically, the IHV of each candidate solution $\mathbf{x}\in\mathcal{P}$ is calculated as:
\begin{equation}
    \mathtt{IHV}(\mathbf{x})=\mathtt{HV}(\mathcal{P})-\mathtt{HV}(\mathcal{P}\setminus\{\mathbf{x}\}),
\end{equation}
where $\mathtt{HV}(\mathcal{P})$ evaluates the Hypervolume of $\mathcal{P}$. Then, the top $\xi$ solutions in $\mathcal{P}$ with the largest IHV are picked up for the expensive function evaluations.


\section{Experimental Setup}
\label{sec:setup}

This section introduces our experimental setup including the benchmark test problems, the peer algorithms along with their parameter settings, and the performance metrics and statistical tests.

\subsection{Benchmark Test Problems}
\label{sec:benchmarks}

In our empirical study, we consider benchmark test problems with disconnect PF segments to constitute our benchmark suite, including ZDT3~\cite{ZitzlerDT00}, DTLZ7~\cite{DebTLZ05} and WFG2~\cite{HubandHBW06} along with their variants dubbed ZDT3$\star$, DTLZ7$\star$ and WFG2$\star$. Their mathematical definitions and characteristics can be found in the supplemental document\footnote{The supplemental document can be found in \url{xxxxxxxxxxxxx}.}. For each benchmark test problem, we set the number of objectives as $m=2$ and the number of variables as $n\in\{3,5,8\}$ respectively in our empirical study~\cite{LiZZL09,LiZLZL09,CaoWKL11,LiKWCR12,LiKCLZS12,LiKWTM13,LiK14,CaoKWL14,LiFKZ14,LiZKLW14,WuKZLWL15,LiKZD15,LiKD15,LiDZ15,LiDZZ17,WuKJLZ17,WuLKZZ17,LiDY18,ChenLY18,ChenLBY18,WuLKZZ19,LiCSY19,Li19,GaoNL19,LiXT19,ZouJYZZL19,LiuLC20,LiXCT20,LiLDMY20,WuLKZ20,LiX0WT20,WangYLK21,LiLLM21,ChenLTL22}.

\subsection{Peer Algorithms and Parameter Settings}
\label{sec:peers}

To validate the competitiveness of our proposed algorithm, we compare its performance with \texttt{ParEGO}~\cite{Knowles06}, \texttt{MOEA/D-EGO}~\cite{ZhangLTV10}, \texttt{K-RVEA}~\cite{ChughJMHS18}, and \texttt{HSMEA}~\cite{HabibSCRM19} widely used in the literature. We do not intend to delineate their working mechanisms here while interested readers are referred to their original papers for details. The parameter settings are listed as follows.
\begin{itemize}
    \item\underline{Number of function evaluations (FEs)}: The initial sample size is set to $11\times n-1$ for all algorithms and the maximum number of FEs is capped as $250$. 
    \item\underline{Reproduction operators}: The parameters associated with the simulated binary crossover~\cite{DebA94} and polynomial mutation~\cite{DebG96} are set as $p_c=1.0$, $\eta_c=20$, $p_m=1/n$, $\eta_m=20$.
    \item\underline{Kriging models}: As for the algorithms that use Kriging for surrogate modeling, the corresponding hyperparameters of the MATLAB Toolbox DACE~\cite{IMM2002-01460} is set to be within the range $[10^{-5},10^5]$.
    \item\underline{Batch size $\xi$}: It is set as $\xi=10$ for our proposed algorithms and $\xi=5$ is set in \texttt{MOEA/D-EGO}.
    \item\underline{Number of repeated runs}: Each algorithm is independently run on each test problem for $31$ times with different random seeds.
\end{itemize}

\subsection{Performance Metric and Statistical Tests}
\label{sec:metrics}

To have a quantitative evaluation of the performance of different algorithms, we use the widely used HV as the performance metric. To have a statistical interpretation of the significance of comparison results, we use the following three statistical measures in our empirical study.
\begin{itemize}
	\item\underline{Wilcoxon signed-rank test}~\cite{Wilcoxon1945IndividualCB}: This is a widely used non-parametric statistical test to conduct a pairwise comparison. In our experiments, we set the significance level as $p=0.05$.
    \item\underline{$A_{12}$ effect size}~\cite{VarghaD00}: This is an effect size measure that evaluates the probability of one algorithm is better than another. Specifically, given a pair of peer algorithms, $A_{12}=0.5$ means they are \textit{equivalent}. $A_{12}>0.5$ denotes that one is better for more than 50\% of the times. $0.56\leq A_{12}<0.64$ indicates a \textit{small} effect size while $0.64 \leq A_{12} < 0.71$ and $A_{12} \geq 0.71$ mean a \textit{medium} and a \textit{large} effect size, respectively. 
    \item\underline{Scott-knott test}~\cite{MittasA13}: This is used to rank the performance of different peer algorithms over $31$ runs on each test problem. In a nutshell, it uses a statistical test and effect size to divide the performance of peer algorithms into several clusters. In particular, the performance of peer algorithms within the same cluster are statistically equivalent. The smaller the rank is, the better performance of the algorithm achieves.
\end{itemize}


\section{Experimental Results }
\label{sec:results}

In this section, our empirical study aims to investigate: 1) the performance of our proposed \our\ compared against the selected peer algorithms; and 2) the effectiveness of the \texttt{MGD-based evolutionary search} and the \texttt{infill criterion} steps of \our.

\subsection{Performance Comparisons with the Peer Algorithms}
\label{sec:peer_comparisons}

The statistical comparison results of the Wilcoxon signed-rank test of the HV values between our proposed \our\ against the other peer algorithms are given in~\pref{tab:RQ1}. From these results, it is clear to see that the HV values obtained by \our\ are statistically significantly better than the other four peer algorithms in all comparisons. As the selected results of the population distribution obtained by different algorithms shown in~\pref{fig:population_distribution}\footnote{Due to the page limit, the complete results are in the supplemental document.}, it is clear to see that the solutions obtained by \our\ not only have a good convergence on the PF, but also have a descent distribution on all disconnected PF segments. In contrast, the other peer algorithms either struggle to converge to the PF or hardly approximate all segments. It is interesting to note that the performance of \texttt{HSMEA} and \texttt{K-RVEA} are acceptable on ZDT3 and DTLZ7$\ast$, which have a relatively small number of disconnected segments, whereas their performance deteriorate significantly when the number of disconnected segments becomes large. 

\begin{figure}[t!]
    \centering
    \includegraphics[width=.7\linewidth]{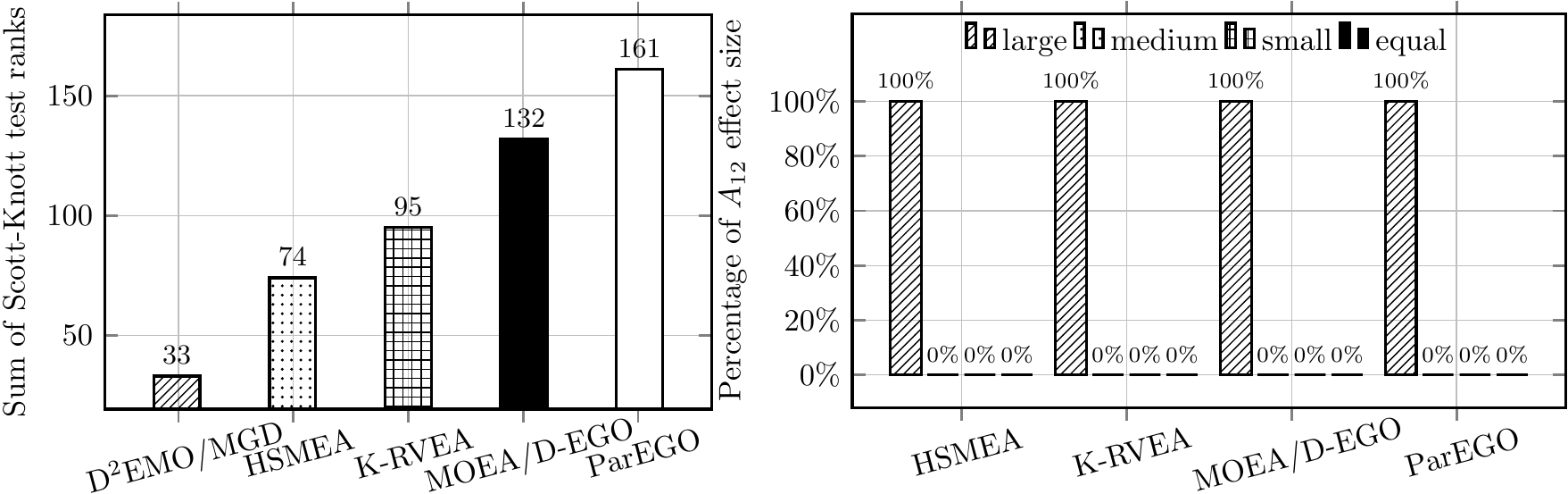}
    \caption{Statistical test results: (a) sum of the Scott-knott test results on all comparisons; (b) percentage of the \textit{equal}, \textit{large}, \textit{medium} and \textit{small} $A_{12}$ effect size, respectively, when comparing \our\ against the other four peer algorithms.}
    \label{fig:statistical_results}
\end{figure}

In addition to the pairwise comparisons, we apply the Scott-knott test to classify their performance into different groups to facilitate a better ranking among these algorithms. Due to the large number of comparisons, it will be messy if we list all ranking results ($11 \times 3=33$ in total). Instead, we pull all the Scott-knott test results together and show their distribution and variance as the bar charts with error bar in~\pref{fig:statistical_results}(a). From this results, we can see that our proposed \our\ is the best algorithm in all comparisons, which confirm the observations from~\pref{tab:RQ1}. In addition, to have a better understanding of the performance difference of \our\ w.r.t. the other peer algorithms, we investigate the comparison results of $A_{12}$ effect size. From the bar charts shown in~\pref{fig:statistical_results}(b), it is clear to see that the better results achieved by \our\ is consistently classified as statistically large.

\begin{table}[t!]
    \centering
	\caption{Comparison results of \texttt{D$^2$EMO/MGD} with the peer algorithms} 
	\label{tab:RQ1}
	\resizebox{\textwidth}{!}{ 
\begin{tabular}{c|c|c|c|c|c|c}
\cline{1-7}
                                              & $n$ & \texttt{D$^2$EMO/MGD} & \texttt{HSMEA}         & \texttt{K-RVEA}        & \texttt{MOEA/D-EGO}    & \texttt{ParEGO}        \\ \hline
\multicolumn{1}{c|}{\multirow{3}{*}{ZDT3}}   & 3   & \bb{1.3199(3.34E-3)}  & 1.2914(5.99E-2)$^\dag$ & 1.1814(1.77E-1)$^\dag$ & 1.1831(9.85E-2)$^\dag$ & 1.1549(4.84E-2)$^\dag$ \\ 
\multicolumn{1}{c|}{}                        & 5   & \bb{1.3271(1.52E-3)}  & 1.2888(7.93E-2)$^\dag$ & 1.1842(7.44E-2)$^\dag$ & 1.0566(1.58E-1)$^\dag$ & 1.0364(1.57E-1)$^\dag$ \\ 
\multicolumn{1}{c|}{}                        & 8   & \bb{1.3260(4.29E-3)}  & 1.1901(1.53E-1)$^\dag$ & 1.2394(1.16E-1)$^\dag$ & 1.0050(1.14E-1)$^\dag$ & 0.7974(1.09E-1)$^\dag$ \\ \hline
\multicolumn{1}{c|}{\multirow{3}{*}{ZDT31}}  & 3   & \bb{1.3813(2.64E-2)}  & 1.3183(2.06E-1)$^\dag$ & 1.0556(2.54E-1)$^\dag$ & 1.0983(2.84E-1)$^\dag$ & 1.0839(9.35E-2)$^\dag$ \\ 
\multicolumn{1}{c|}{}                        & 5   & \bb{1.3706(2.74E-2)}  & 1.3064(6.99E-2)$^\dag$ & 1.1107(1.72E-1)$^\dag$ & 1.0351(2.13E-1)$^\dag$ & 1.0068(8.73E-2)$^\dag$ \\ 
\multicolumn{1}{c|}{}                        & 8   & \bb{1.3655(7.06E-2)}  & 1.1851(1.14E-1)$^\dag$ & 1.1098(1.65E-1)$^\dag$ & 0.9112(1.21E-1)$^\dag$ & 0.8659(2.16E-1)$^\dag$ \\ \hline
\multicolumn{1}{c|}{\multirow{3}{*}{ZDT32}}  & 3   & \bb{1.2818(2.18E-2)}  & 1.2309(9.73E-2)$^\dag$ & 1.1852(7.96E-2)$^\dag$ & 1.1265(4.74E-2)$^\dag$ & 1.1170(1.52E-2)$^\dag$ \\ 
\multicolumn{1}{c|}{}                        & 5   & \bb{1.2610(8.73E-2)}  & 1.1801(9.62E-2)$^\dag$ & 1.1359(6.86E-2)$^\dag$ & 1.0960(6.78E-2)$^\dag$ & 1.0801(1.07E-1)$^\dag$ \\ 
\multicolumn{1}{c|}{}                        & 8   & \bb{1.2860(2.12E-2)}  & 1.1604(9.41E-2)$^\dag$ & 1.1550(8.02E-2)$^\dag$ & 1.0355(1.04E-1)$^\dag$ & 1.0175(1.15E-1)$^\dag$ \\ \hline
\multicolumn{1}{c|}{\multirow{3}{*}{ZDT33}}  & 3   & \bb{0.9089(6.31E-4)}  & 0.8643(5.41E-2)$^\dag$ & 0.8683(7.07E-2)$^\dag$ & 0.8578(3.22E-2)$^\dag$ & 0.8249(2.46E-2)$^\dag$ \\ 
\multicolumn{1}{c|}{}                        & 5   & \bb{0.9075(1.04E-3)}  & 0.8508(3.32E-2)$^\dag$ & 0.8845(6.85E-2)$^\dag$ & 0.8148(1.83E-2)$^\dag$ & 0.7487(5.64E-2)$^\dag$ \\ 
\multicolumn{1}{c|}{}                        & 8   & \bb{0.9036(7.43E-2)}  & 0.8130(9.20E-2)$^\dag$ & 0.8630(6.48E-2)$^\dag$ & 0.7911(3.25E-2)$^\dag$ & 0.7126(4.45E-2)$^\dag$ \\ \hline \hline
\multicolumn{1}{c|}{\multirow{3}{*}{WFG2}}   & 3   & \bb{5.9298(3.99E-2)}  & 5.7944(2.10E-1)$^\dag$ & 5.7083(1.48E-1)$^\dag$ & 4.9327(3.51E-1)$^\dag$ & 4.6709(2.84E-1)$^\dag$ \\ 
\multicolumn{1}{c|}{}                        & 5   & \bb{5.9799(3.14E-2)}  & 5.8221(1.11E-1)$^\dag$ & 5.4672(1.40E-1)$^\dag$ & 4.6521(4.99E-1)$^\dag$ & 4.5911(6.76E-1)$^\dag$ \\ 
\multicolumn{1}{c|}{}                        & 8   & \bb{5.7457(6.63E-2)}  & 5.1535(3.36E-1)$^\dag$ & 5.1191(3.18E-1)$^\dag$ & 4.0520(4.96E-1)$^\dag$ & 3.8378(4.09E-1)$^\dag$ \\ \hline
\multicolumn{1}{c|}{\multirow{3}{*}{WFG21}}  & 3   & \bb{6.2155(4.87E-2)}  & 5.5538(4.16E-1)$^\dag$ & 5.5407(3.49E-1)$^\dag$ & 4.9540(3.22E-1)$^\dag$ & 4.8377(2.97E-1)$^\dag$ \\ 
\multicolumn{1}{c|}{}                        & 5   & \bb{6.2630(2.37E-2)}  & 5.0867(5.17E-1)$^\dag$ & 5.5867(9.45E-2)$^\dag$ & 4.7417(4.15E-1)$^\dag$ & 4.6174(3.08E-1)$^\dag$ \\ 
\multicolumn{1}{c|}{}                        & 8   & \bb{6.0309(5.79E-2)}  & 4.5420(3.14E-1)$^\dag$ & 5.2068(3.06E-1)$^\dag$ & 4.0720(3.10E-1)$^\dag$ & 3.9503(4.48E-1)$^\dag$ \\ \hline
\multicolumn{1}{c|}{\multirow{3}{*}{WFG22}}  & 3   & \bb{2.8400(6.29E-2)}  & 2.8154(1.42E-1)$^\dag$ & 2.5958(1.55E-1)$^\dag$ & 2.0903(1.65E-1)$^\dag$ & 2.0581(1.68E-1)$^\dag$ \\ 
\multicolumn{1}{c|}{}                        & 5   & \bb{2.8638(6.03E-2)}  & 2.6386(1.21E-1)$^\dag$ & 2.3946(6.85E-2)$^\dag$ & 1.8671(1.47E-1)$^\dag$ & 1.9506(1.78E-1)$^\dag$ \\ 
\multicolumn{1}{c|}{}                        & 8   & \bb{2.6378(1.14E-1)}  & 2.0694(2.54E-1)$^\dag$ & 1.8419(1.50E-1)$^\dag$ & 1.3691(5.16E-1)$^\dag$ & 1.3248(2.54E-1)$^\dag$ \\ \hline
\multicolumn{1}{c|}{\multirow{3}{*}{WFG23}}  & 3   & \bb{5.7030(3.45E-1)}  & 5.6479(4.02E-1)$^\dag$ & 5.4574(2.62E-1)$^\dag$ & 4.9859(2.54E-1)$^\dag$ & 5.0298(4.56E-1)$^\dag$ \\ 
\multicolumn{1}{c|}{}                        & 5   & \bb{5.9445(9.55E-2)}  & 5.3802(3.19E-1)$^\dag$ & 5.4525(1.42E-1)$^\dag$ & 4.9615(2.60E-1)$^\dag$ & 4.7642(5.69E-1)$^\dag$ \\ 
\multicolumn{1}{c|}{}                        & 8   & \bb{5.6221(9.38E-2)}  & 4.7431(2.40E-1)$^\dag$ & 4.8954(2.38E-1)$^\dag$ & 4.2413(3.71E-1)$^\dag$ & 4.0819(2.20E-1)$^\dag$ \\ \hline  \hline
\multicolumn{1}{c|}{\multirow{3}{*}{DTLZ7}}  & 3   & \bb{1.3234(7.52E-3)}  & 1.3195(1.84E-2)$^\dag$ & 1.2407(1.18E-2)$^\dag$ & 1.2550(2.56E-2)$^\dag$ & 1.2076(6.05E-2)$^\dag$ \\ 
\multicolumn{1}{c|}{}                        & 5   & \bb{1.3330(1.29E-3)}  & 1.3138(1.73E-2)$^\dag$ & 1.2371(2.14E-2)$^\dag$ & 1.2131(1.20E-1)$^\dag$ & 1.1303(5.55E-2)$^\dag$ \\ 
\multicolumn{1}{c|}{}                        & 8   & \bb{1.3297(2.99E-3)}  & 1.3123(1.09E-2)$^\dag$ & 1.2263(2.52E-2)$^\dag$ & 1.1591(1.13E-1)$^\dag$ & 1.0667(1.21E-1)$^\dag$ \\ \hline
\multicolumn{1}{c|}{\multirow{3}{*}{DTLZ71}} & 3   & \bb{1.3996(2.64E-3)}  & 1.3959(1.02E-2)$^\dag$ & 1.3289(1.79E-2)$^\dag$ & 1.2916(7.39E-1)$^\dag$ & 1.2735(2.84E-2)$^\dag$ \\ 
\multicolumn{1}{c|}{}                        & 5   & \bb{1.4040(2.96E-3)}  & 1.3936(9.43E-3)$^\dag$ & 1.3190(1.26E-2)$^\dag$ & 1.2746(3.82E-1)$^\dag$ & 1.0609(1.67E-1)$^\dag$ \\ 
\multicolumn{1}{c|}{}                        & 8   & \bb{1.4034(4.42E-3)}  & 1.3762(1.07E-1)$^\dag$ & 1.3146(1.50E-2)$^\dag$ & 1.0508(1.41E-1)$^\dag$ & 0.9420(2.51E-1)$^\dag$ \\ \hline
\multicolumn{1}{c|}{\multirow{3}{*}{DTLZ72}} & 3   & \bb{1.4013(9.67E-3)}  & 1.4013(1.69E-2)$^\dag$ & 1.3292(2.58E-2)$^\dag$ & 1.3006(1.44E-1)$^\dag$ & 1.2863(4.18E-2)$^\dag$ \\ 
\multicolumn{1}{c|}{}                        & 5   & \bb{1.4075(2.62E-3)}  & 1.3926(1.00E-2)$^\dag$ & 1.3313(1.72E-2)$^\dag$ & 1.1908(1.14E-1)$^\dag$ & 1.2150(6.59E-2)$^\dag$ \\ 
\multicolumn{1}{c|}{}                        & 8   & \bb{1.4052(2.34E-3)}  & 1.3875(1.20E-1)$^\dag$ & 1.3170(1.53E-2)$^\dag$ & 1.0411(8.06E-2)$^\dag$ & 0.9540(2.36E-1)$^\dag$ \\ \hline
\end{tabular}
}
\end{table}

\begin{figure}[t!]
    \centering
    \includegraphics[width=\linewidth]{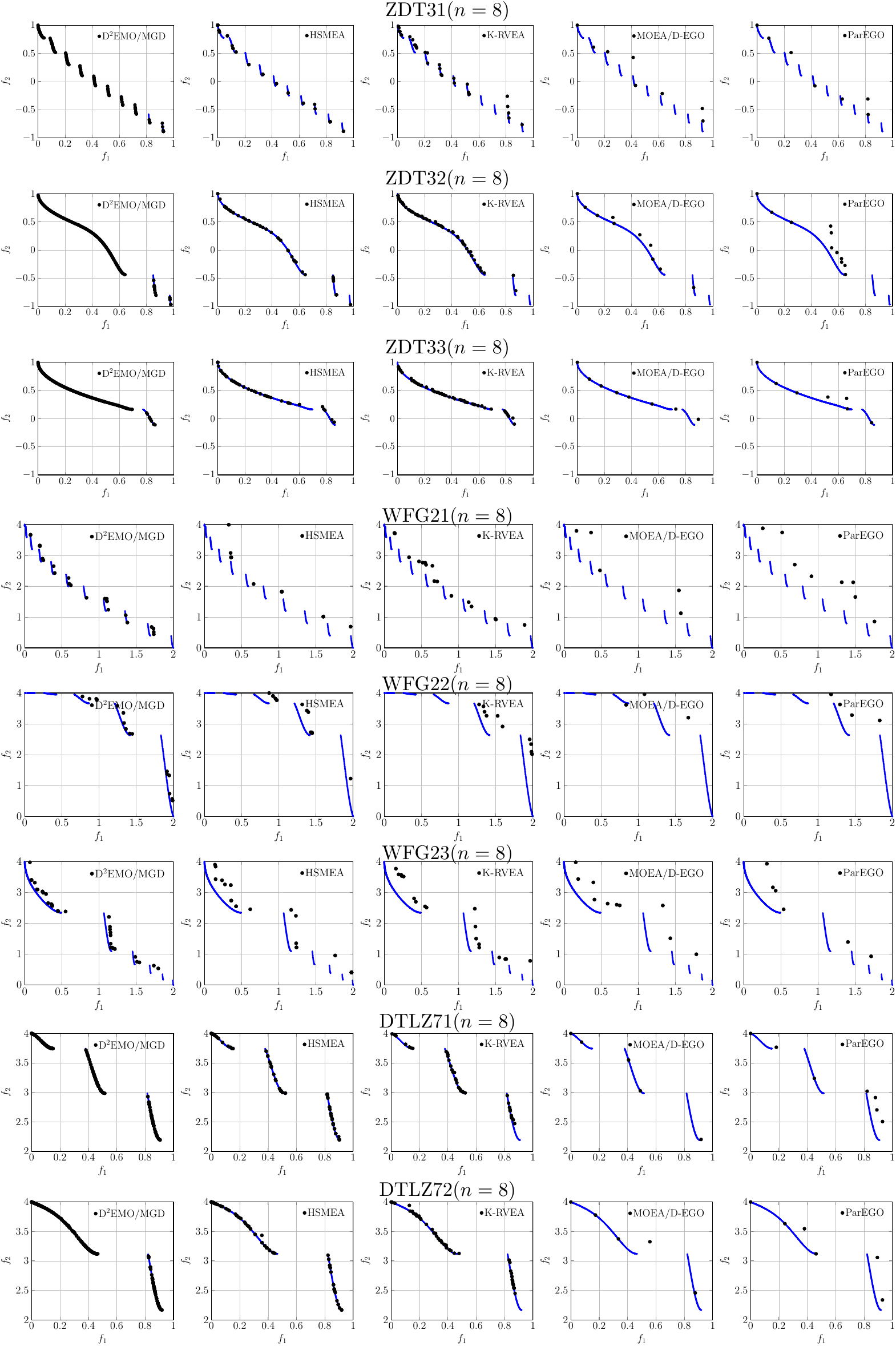}
    \caption{Non-dominated solutions obtained by different algorithms (with the medium HV values) on ZDT3$\ast$, WFG2$\ast$, and DTLZ7$\ast$ ($n=8$), respectively.}
    \label{fig:population_distribution}
\end{figure}

\subsection{Ablation Study}
\label{sec:ablation}

\begin{figure}[t!]
    \centering
    \includegraphics[width=.7\linewidth]{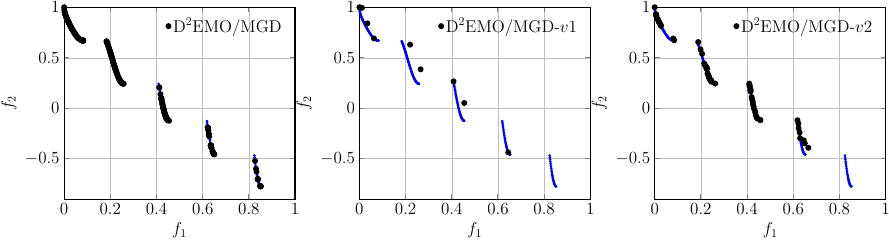}
    \caption{Non-dominated solutions (with the medium HV value) found by different variants of \our\ on ZDT3 with $n=8$.}
    \label{fig:rq2_pop}
\end{figure}

\vspace{-0.5em}
In this subsection, we empirically investigate the effectiveness of two key algorithmic components of \our. More specifically, we first compare the performance of \our\ w.r.t. the variant \our-$v1$ without using the \texttt{MGD-based evolutionary search} step. Instead, it applies the widely used simulated binary crossover~\cite{DebA94} as the alternative operator for offspring reproduction. From the selected results shown in~\pref{fig:rq2_pop}, we can see that \our-$v1$ can only find a very limited number of solutions on the PF with a poor diversity. Thereafter, we compare the performance of \our\ w.r.t. the variant \our-$v2$ without using the \texttt{infill criterion} introduced in~\pref{sec:infill}. In particular, it uses a random selection to recommend the candidates for expensive function evaluations. From the selected results shown in~\pref{fig:rq2_pop}, we can see that \our-$v2$ can only find some of the disconnected PF segments.

\section{Conclusion}
\label{sec:conclusions}

Most, if not all, existing data-driven EMO algorithms, directly apply evolutionary operators for offspring reproduction, while the regularity information embedded in the surrogate models has been unfortunately ignored. Bearing this consideration in mind, this paper, for the first time, investigates the use of MGD to leverage the latent information of the surrogate models to accelerate the convergence of the evolutionary population towards the PF. Due to the extra diversity provided by the exploration along the approximated PS manifold, our proposed \our\ has shown strong performance on the selected benchmark test problems with disconnected PF segments. In addition, the ablation study also confirms the usefulness of the batched infill criterion guided by the IHV. Since this paper only considers problems with two objectives, its first potential extension is the scalability for problems with many objectives. In addition, it is well known that a biased mapping between the decision variables and the objective functions can raise significant challenges in EMO~\cite{Deb99}. This will be aggravated when using the MGD, since the gradients along the approximated PS manifold can be largely biased towards a local niche. In addition, it is also interesting to investigate the performance of data-driven EMO algorithms for constrained multi-objective optimization problems~\cite{LiCFY19,BillingsleyLMMG19,BillingsleyLMMG21,ShanL21,YangHL21}.

\section*{Acknowledgment}
This work was supported by UKRI Future Leaders Fellowship (MR/S017062/1), EPSRC (2404317), NSFC (62076056), Royal Society (IES/R2/212077) and Amazon Research Award.

\bibliographystyle{IEEEtran}
\bibliography{IEEEabrv,mgd}

\end{document}